\documentclass[journal]{IEEEtran}

\usepackage{xcolor,soul,framed} 

\colorlet{shadecolor}{yellow}
\usepackage[pdftex]{graphicx}
\graphicspath{{../pdf/}{../jpeg/}}
\DeclareGraphicsExtensions{.pdf,.jpeg,.png}

\usepackage[cmex10]{amsmath}
\usepackage{array}
\usepackage{mdwmath}
\usepackage{mdwtab}
\usepackage{eqparbox}
\usepackage{url}

\usepackage{amssymb}
\usepackage{multirow}
\newcommand{\RNum}[1]{\uppercase\expandafter{\romannumeral #1\relax}}
\usepackage[ruled]{algorithm2e}
\usepackage{hyperref}
\hypersetup{
    colorlinks=true,            
    linkcolor=blue,             
    filecolor=cyan,             
    urlcolor=magenta,           
    pdftitle={Overleaf Example},
    pdfpagemode=FullScreen,
    }
    
\begin{document}
\bstctlcite{IEEEexample:BSTcontrol}
    \title{Continuous Sign Language Recognition Based on Motor attention mechanism and frame-level Self-distillation}
  \author{
   \IEEEauthorblockN{Qidan~Zhu$^{1}$, 
    Jing~Li$^{1*}$, 
    Fei~Yuan$^2$, 
    Quan~Gan$^1$}

    \IEEEauthorblockA{$^1$ College of Intelligent Systems Science and Engineering, Harbin Engineering University, Harbin, 150001, China}
    \IEEEauthorblockA{$^2$ Northwest Institute of Mechanical and Electrical Engineering, Xianyang, 712099, China}

 \thanks{*Corresponding author\par
Email addresses: \par
zhuqidan@hrbeu.edu.cn (Qidan Zhu), \par
ljing@hrbeu.edu.cn (Jing Li), \par
bohelion@hrbeu.edu.cn (Fei Yuan), \par
gquan@hrbeu.edu.cn (Quan Gan)}
  }

\maketitle

\begin{abstract}
Changes in facial expression, head movement, body movement and gesture movement are remarkable cues in sign language recognition, and most of the current continuous sign language recognition(CSLR) research methods mainly focus on static images in video sequences at the frame-level feature extraction stage, while ignoring the dynamic changes in the images. In this paper, we propose a novel motor attention mechanism to capture the distorted changes in local motion regions during sign language expression, and obtain a dynamic representation of image changes. And for the first time, we apply the self-distillation method to frame-level feature extraction for continuous sign language, which improves the feature expression without increasing the computational resources by self-distilling the features of adjacent stages and using the higher-order features as teachers to guide the lower-order features. The combination of the two constitutes our proposed holistic model of CSLR Based on motor attention mechanism and frame-level Self-Distillation (MAM-FSD), which improves the inference ability and robustness of the model. We conduct experiments on three publicly available datasets, and the experimental results show that our proposed method can effectively extract the sign language motion information in videos, improve the accuracy of CSLR and reach the state-of-the-art level.
\end{abstract}

\begin{IEEEkeywords}
 continuous sign language recognition; frame-level self-distillation; motor attention mechanism; dynamic feature representation
\end{IEEEkeywords}

%
\IEEEpeerreviewmaketitle


\section{Introduction}

\IEEEPARstart{A}{s} a basic tool for communication with deaf people, it takes a long time and effort to learn and master sign language, which hinders the communication between normal people and them. CSLR has considerable social value and significance as it can realize communication with deaf people in real life by translating the dynamic video of sign language into understandable sentences.\par

For CSLR, researchers have mainly proceeded from the joint frame-level feature extraction and temporal feature extraction of sign language videos. In \cite{hu2023continuous}\cite{xie2023multi}, continuous sign language feature extraction is performed based on the sequential order, and methods such as frame-level feature correlation module, multi-scale local temporal similarity fusion are proposed for extracting the two types of features, respectively. And continuous sign language features are directly extracted for recognition by the proposed dual-stream visual coder method in \cite{chen2022two}. The loss function is also focused on CTC loss\cite{li2020reinterpreting}, multilevel CTC loss\cite{zhu2022multi}. With the rapid development of deep learning, multiple cues, attention mechanisms, etc., which have excellent performance in applications such as image classification, target detection and semantic segmentation, have also been gradually applied to CSLR\cite{zhou2020spatial}\cite{huang2018attention}\cite{habili2004segmentation}\cite{zhang2023multimodal}. In these methods, whether it is 2DCNN or 3DCNN, the focus is mainly placed in the separate whole image frames in the video. It is worth noting that what is crucial in sign language recognition is the changes in facial expression, head movement, body movement and gesture movement, which serve as significant clues for sign language recognition. Figure 1 shows the motion heat map obtained by the inter-frame difference method and it can be seen that for consecutive complete image frames, the regions of change are concentrated in the above mentioned regions.\par

\begin{figure*}
  \begin{center}
  \includegraphics[width=3.5in]{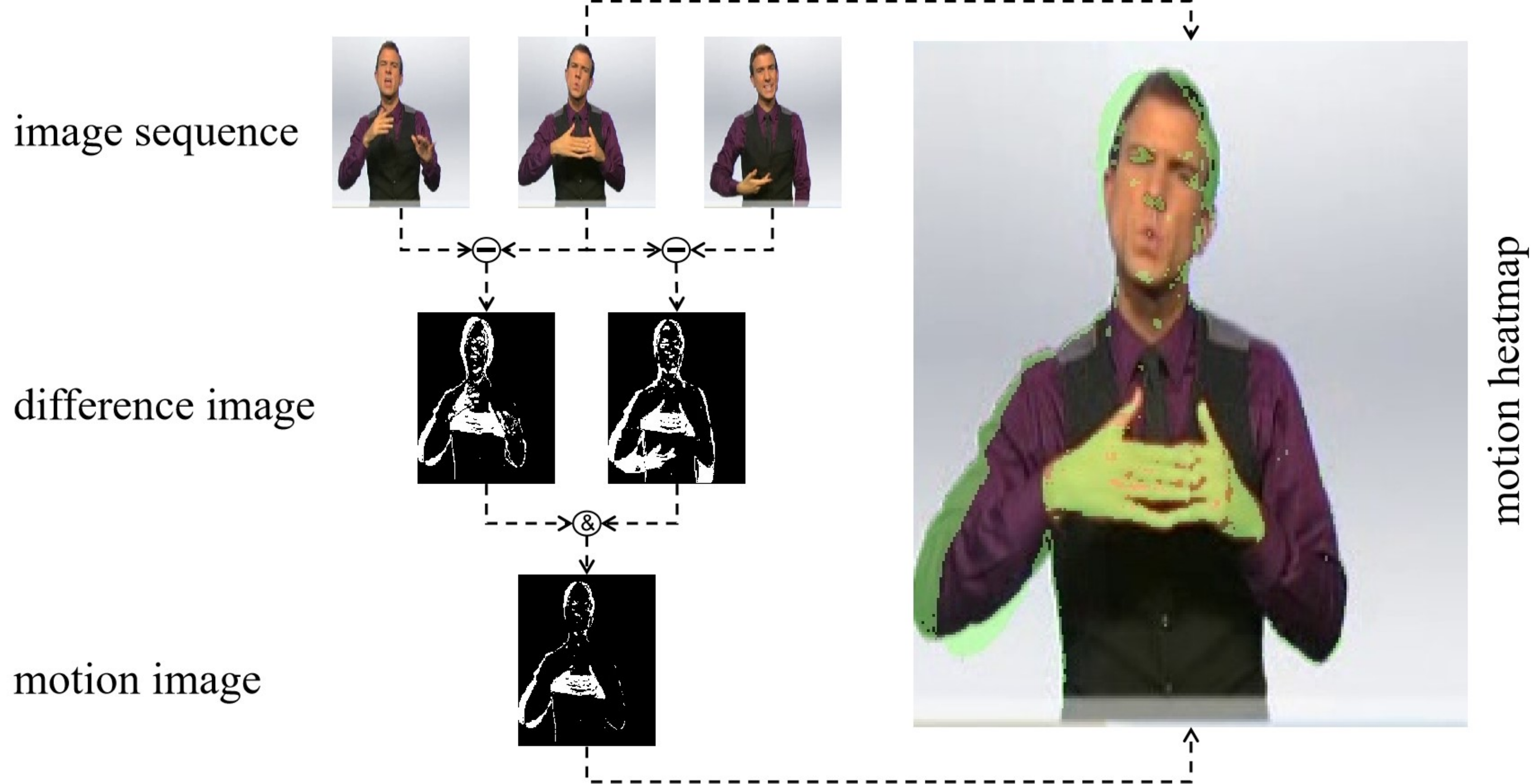}\\
  \caption{The motion heat map obtained by the inter-frame difference method.}\label{fig:ljxy1}
  \end{center}
\end{figure*}

The core purpose of this paper is to study the dynamic changes between frames, obtain a dynamic expression of image changes, capture distorted changes in local motion regions when generating sign language expressions, and improve the accuracy of sign language recognition. Reference \cite{hu2023continuous} performs continuous sign language recognition through a correlation module that captures the trajectory between neighboring frames and a recognition module that locates the body region, and its idea of studying the dynamic changes between frames is similar to the research work in this paper, with the difference that it goes for capturing the correlation of the local regions more. The attention model of Transfomer and its variants in CSLR\cite{du2022full}\cite{yin2020better}\cite{hu2023signbert+}\cite{zhao2021conditional} has recently performed well and achieved success. These methods are more focused on improving the distribution of temporal features in the global region to enhance beneficial features. This paper aims to make the model intuitively focus on changing regions and proposes a new motor attention mechanism to enhance motion information in the video. In addition, unlike other methods for ordinary processing of frame-level features, this paper also proposes a frame-level feature self-distillation method. At present, some methods in CSLR use the self-distillation method in temporal feature extraction\cite{chen2022two}\cite{min2021visual}. This paper applies the self-distillation method to frame-level feature extraction of continuous sign language, improving feature expression without increasing computational resources, thereby improving model performance and robustness.\par

Specifically, the model MAM-FSD proposed in this paper mainly consists of motor attention mechanism module and frame-level self-distillation method. The proposed motor attention mechanism module uses a multi-layer 3D convolution to perform a weighted summation of the pixels corresponding to adjacent frames of consecutive sign language videos, and the result of the summation is normalized and multiplied with the original feature map to enhance the dynamic motion information. This process does not use global pooling to highlight the target dimension as in spatial, channel attention mechanisms \cite{hu2018squeeze}\cite{woo2018cbam}. And unlike \cite{lee2018motion}\cite{zhang2023extracting} that use adjacent frame feature subtraction and its variants for feature enhancement, these methods are often used for motion information extraction. The proposed frame-level self-distillation method trains higher dimensional features as teachers and lower dimensional features as students for frame-level features with different dimensions, enabling better learning and updating of frame-level features. Encouragingly, the model achieves new state-of-the-art accuracy on three large-scale publicly available datasets RWTH\cite{koller2015continuous}, RWTH-T\cite{camgoz2018neural}, and CSL-Daily\cite{zhou2021improving}, under RGB inputs only, which is closely related to our new approach that focuses more on continuous frame dynamics and better utilizes frame-level features.  The superiority of the model is demonstrated by a comprehensive comparison with other advanced methods. The visualization results validate the impact of the model on motion change information.\par

In summary, the main contributions of this paper are as follows:\par

\begin{itemize}
\item[$\bullet$] A new motor attention mechanism is proposed to improve the model's inference ability by paying more attention to the dynamic changes in consecutive frames of the video.
\item[$\bullet$] A frame-level self-distillation method is proposed, which enhances the ability to express features without increasing computational resources by instructing students to operate on the frame-level features of consecutive video frames.
\item[$\bullet$] A CSLR model MAM-FSD is proposed to achieve new state-of-the-art accuracy in CSLR on three large-scale datasets RWTH, RWTH-T, and CSL-Daily, under the condition of RGB input only.
\end{itemize}

\begin{figure*}
  \begin{center}
  \includegraphics[width=6.0in,height=2.0in]{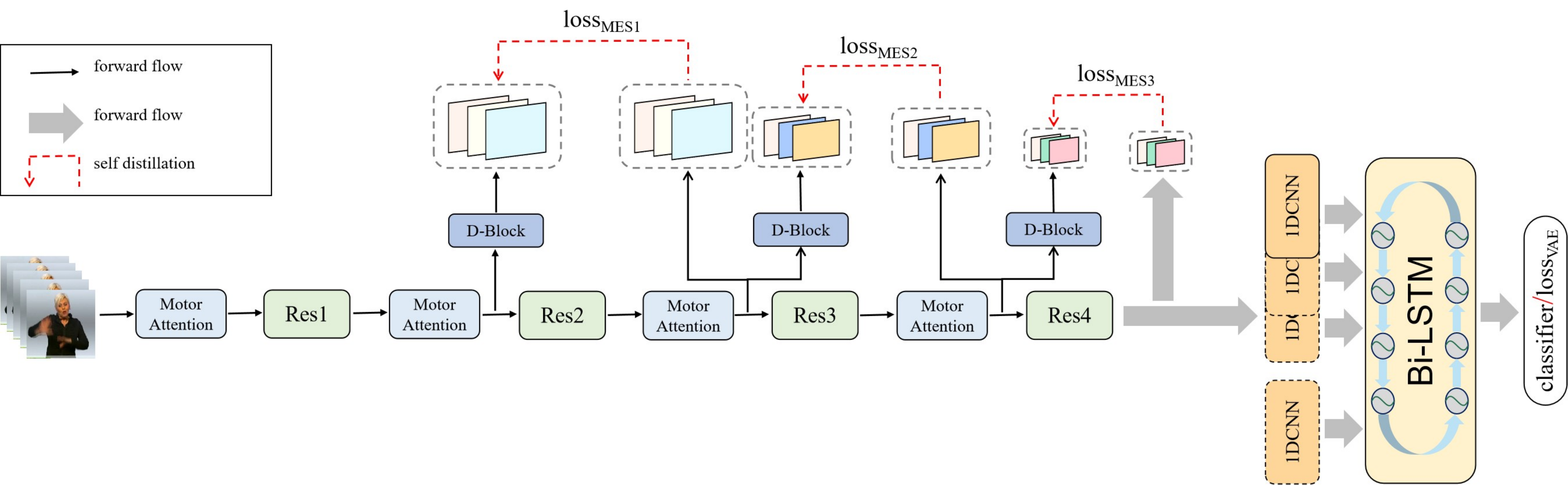}\\
  \caption{Overview of the new MAM-FSD. Firstly, CNN is used to capture frame-level features, followed by 1DCNN+BiLSTM for temporal modeling, and finally, a classifier is used to predict sentences. We place the proposed motor attention mechanism module and frame-level self-distillation method in the frame-level feature extraction section.}\label{fig:ljxy2}
  \end{center}
\end{figure*}

\section{Related Work}

\subsection {Continuous Sign Language Recognition}

CSLR exists as a weakly supervised approach problem due to the lack of strict correspondence between video frames and labeled sequences. Initially, the recognition results were inferred by a combination of manually designed features and traditional machine learning such as HMM and DTW\cite{talukdar2022vision}\cite{zhang2014threshold}\cite{wang2014similarity}. With the development of deep learning and the widespread use of CNNs, CSLR has also grown tremendously. The CSLR model composed of networks CNN+HMM\cite{koller2018deep} and CNN+LSTM+HMM\cite{koller2019weakly} combined with traditional machine learning algorithms, as well as networks CNN+RNN\cite{gao2021rnn}, CNN+LSTM\cite{guo2019hierarchical}, and CNN+BiLSTM\cite{wei2020semantic} combined with various neural networks, not only achieves success but also forms a classic and solid foundation in the research of continuous sign language recognition. Convolution of various dimensions has also been used by researchers in their development. It is important to mention that the common problem of all convolutional neural networks is that they only focus on the local features of the whole image. We take an alternative approach to recognize the sign language by focusing more on the changing parts of each image frame to get the dynamics between consecutive frames and thus recognize the sign language.There are also studies that guide the model by focusing on the hands and facial regions of the presenter\cite{xiao2020multi}\cite{papadimitriou2020multimodal}, using pre-extracted gesture key points as supervision. Our approach allows the model to learn noteworthy motion change regions end-to-end and self-learning the connections between consecutive frames, without relying on additional clues such as pre-extracted body keypoints\cite{zhou2020spatial} or multiple streams\cite{cui2019deep}, which requires more computation to utilize hand and face information. We aim to obtain an efficient model for CSLR features.\par

\subsection {Attention-related methods}

Attention-related methods have been widely applied in current research in video recognition, promoting the development of video recognition\cite{fu2021learning}\cite{zhao2022alignment}. The Transfomer model and its variants\cite{du2022full}\cite{yin2020better} are applied in CSLR. Hu et al.\cite{hu2022transformer} proposed a keyframe extraction algorithm for continuous sign language videos, which utilizes image differentiation and image blur detection techniques to adaptively calculate the difference threshold and extract keyframes from symbol videos. Zhao et al.\cite{hu2023signbert+} proposed the first self-supervised pre-training framework, SignBERT+, which includes manual pre-testing, using both hands as visual gestures and carefully embedding the pose state and spatial-temporal position information of each visual gesture. The research directions of the same idea \cite{zhang2023extracting}\cite{jiang2019stm} are all based on the method of studying the dynamic changes between frames. Among them, only \cite{hu2023continuous} is used in CSLR through a correlation module that captures trajectories between adjacent frames and a recognition module that locates body regions. The method in \cite{diba2018spatio} also performs well and is applied in action classification by introducing a new "spatial-temporal channel correlation" (STC) block to simulate the correlation between 3DCNN channels. Similarly, in \cite{wang2020video}, Wang et al. proposed a learnable correlation operator, which is a frame-to-frame matching on convolutional feature maps in different layers of the network. These methods are more focused on capturing the correlation of local regions. This paper aims to make the model intuitively focus on changing regions and proposes a new motor attention mechanism to enhance motion information in videos.\par

\subsection {Self-distillation method}

Knowledge distillation is a common approach for model compression\cite{buciluǎ2006model}\cite{hinton2015distilling}, and models consisting of three common knowledge types Response-Based Knowledge, Feature-Based Knowledge, and Relation-Based Knowledge have also been applied to CSLR\cite{min2021visual}\cite{zhu2023continuous}. Attentional distillation and self-distillation, which have since expanded and emerged, have also shown great success recently. Early attention distillation\cite{komodakis2017paying} established attention distillation loss in the corresponding stages of teacher and student models, and its attention feature map construction method was Activation based or Gradient based. \cite{hou2019learning} made the first attempt to use the network's own attention map as the distillation target, and proposed a self attention distillation method (Self Attention Distillation (SAD)), SAD allows a network to utilize the attention maps obtained from its own layers as distillation targets for its lower layers and obtain substantial improvements without any additional supervision or labeling. Subsequently, in applications that require intolerable errors, such as autonomous driving and medical image analysis, where further improvements in prediction and analysis accuracy are needed, along with shorter response times, the self-distillation method arose. \cite{zhang2019your} provided a single neural network executable program of varying depths that allows for adaptive accuracy and efficiency trade-offs on resource-limited edge devices, significantly improving the accuracy of the algorithms while introducing virtually no computational cost. \cite{ji2021refine} proposed feature self-distillation methods that can be utilized for self knowledge distillation using soft labeling and feature map distillation. In this paper, the self-distillation method is applied to frame-level feature extraction for continuous sign language for the first time to improve the feature representation without increasing computational resources, which in turn improves the model performance and robustness.\par

\section{Methodology}

The backbone of the MAM-FSD model consists of a frame-level feature extractor, a 1D CNN+BiLSTM temporal feature extractor, and a CTC classifier to perform predictions, using the VAC model\cite{min2021visual} as the base model, as shown in Figure 2. The motor attention mechanism module and the frame-level self-distillation method are placed in the frame-level feature extraction section to identify dynamic information and better train the model parameters. For a given input of a continuous sign language video $V=(x_1,x_2,...,x_T)=\{{x_t|_1^T\in \mathbb{R}^{T\times c\times h\times w}}\}$ containing frame $T$, convert it into comprehensible sentences. Specifically, the frame-level feature extractor first processes the input frames into frame-level features $f_{spatial} = F_s(V)\in \mathbb{R}^{T\times c^{'}\times h^{'}\times w^{'}}$. Then, 1D CNN and BiLSTM  perform short-term and long-term temporal modeling based on these extracted visual representations respectively, to obtain temporal features $f_{temporal} = F_t(f_{spatial})\in \mathbb{R}^{T^{'}\times c^{'}\times h^{'}\times w^{'}}$, where $F_t$ is the temporal feature extractor. Finally, a fully connected layer combined with softmax is used to predict the probability of the target sequence, and the overall model is trained using the CTC loss function pairs.\par

\begin{figure*}
  \begin{center}
  \includegraphics[width=2.0in,height=3.0in]{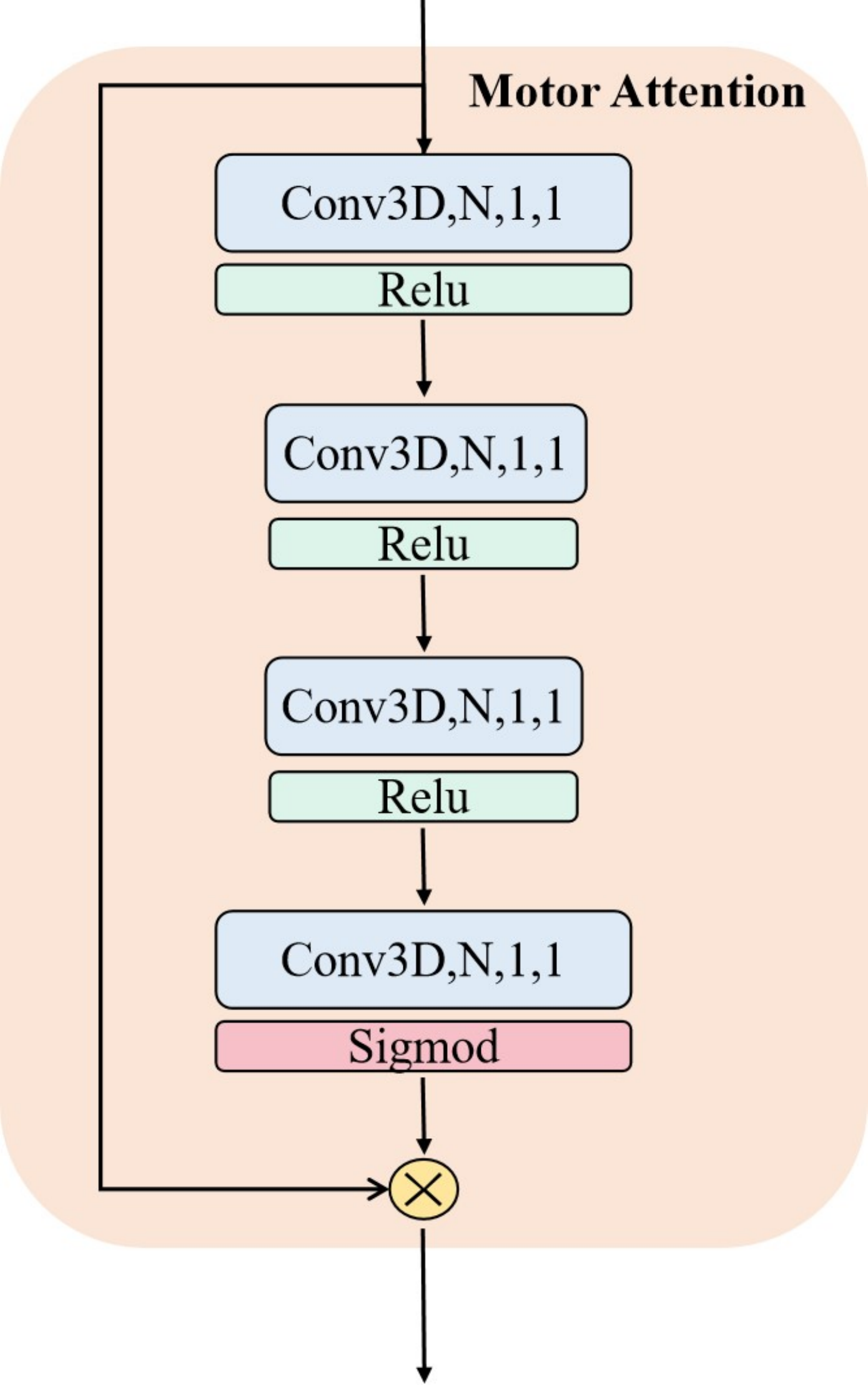}\\
  \caption{Structure diagram of motor attention mechanism.}\label{fig:ljxy3}
  \end{center}
\end{figure*}

\subsection {Motor attention mechanism}

Crucial in sign language recognition are the changes in human expression, head movement, body movement and gesture movement, which are significant cues foe sign language recognition. We propose the motor attention mechanism to get the dynamic expression of image changes by studying the dynamic changes between frames, capturing the distortion changes of local motion regions when producing sign language expressions, and thereby improving the accuracy of sign language recognition.\par

The motor attention mechanism module uses multi-layer 3D convolution to weighted sum the corresponding pixels of adjacent frames in continuous sign language videos, normalizes the sum results, and multiplies them with the original feature map to enhance dynamic motion information. The details are shown in Figure 3.\par

For the feature map $F_{input}=(f_1,f_2,...,f_T)=\{{f_t|_1^T\in \mathbb{R}^{C\times T\times H\times W}}\}$ with an input timing length of $T$, where $f_t$ is the t-th feature map, $H\times W$ is the size of $f_t$, and $C$ is the number of channels. Firstly, a 3D convolution with a kernel $N\times 1\times 1$ is performed, and the calculation process is as follows:\par

\begin{equation}
f_{motor} = \sum_{\substack{t=1}}^{\substack{T}}\sum_{\substack{i=1}}^{\substack{H}}\sum_{\substack{j=1}}^{\substack{W}}\sum_{\substack{c=1}}^{\substack{C}}\sum_{\substack{n=-m}}^{\substack{m}}(f_{t,i,j}^c(t+n)\cdot w^c(m+n))
\end{equation}

Among them, $f_{motor}$ represents the dynamic feature information after convolution calculation, $m=\lfloor \frac{N}{2} \rfloor$, and $f_{t,i,j}^c(t+n)$ represents traversing $N$ neighborhood feature maps with $t$ as the center and calculating pixel by pixel. The computation is completed after the activation function $Relu$, which has the following expression:\par

\begin{equation}
f_{motor}^{'} = Relu(f_{motor})\in \mathbb{R}^{C^{'}\times T\times H\times W}
\end{equation}

\noindent where $C^{'}$ is the size of the number of channels after feature extraction. The same calculation is performed twice more according to the method of formulas (1) and (2). In order to transform dynamic information into intensity distribution, the eigenvalues are constrained to the range of 0-1, and the activation function is set to $Sigmod$. The last layer of feature expression is as follows:\par

\begin{equation}
f_{motor}^{p} = Sigmod(Conv3D(f_{motor}))\in \mathbb{R}^{C\times T\times H\times W}
\end{equation}

\noindent where $f_{motor}^{p}$ is the final dynamic information intensity distribution feature map, at which time the number of channels is restored to the input size $C$. Then the final output $F_out$ is:\par

\begin{equation}
F_{out} = f_{motor}^{p}\cdot F_{in}\in \mathbb{R}^{C\times T\times H\times W}
\end{equation}

\subsection {Frame-level self-distillation}

For frame-level features, we propose frame-level self-distillation technology and construct a frame-level self-distillation framework. Firstly, the convolutional neural network that constitutes the target frame-level feature extractor is divided into several stages based on its depth and original structure. In this paper, we use ResNet34\cite{he2016deep} as the frame-level feature extraction backbone, which is divided into four parts according to the Res block. Secondly, after the first three stages, a D-block is set, which is a 2D convolution. By setting channel and step parameters, the channel of the feature map is dimensionally increased and the size of the feature map is reduced to match the size of the higher dimensional feature map.\par

\begin{table*}[!htbp]
\centering
\caption{we compare the performance with other state-of-the-art methods on RWTH, RWTH-T, and CSL-Daily, where Full indicates that only the full RGB image is used for recognition, and Extra clues indicates that other cues are used for recognition($\surd$ indicates that they are used, and - indicates that they are not used)}
\label{tab:aStrangeTable1}
\begin{tabular}{c|c|c|c|c|c|c|c|c}
\hline  
\multirow{2}{*}{Methods}& \multirow{2}{*}{Full}& \multirow{2}{*}{Extra clues}& \multicolumn{2}{c|}{RWTH}& \multicolumn{2}{c|}{RWTH-T}& \multicolumn{2}{c}{CSL-Daily}\\
\cline{4-9}
 & & & Dev(\%)& Test(\%)& Dev(\%)& Test(\%)& Dev(\%)& Test(\%)\\
\hline
LS-HAN\cite{huang2018video}& -& $\surd$& -& -& -& -& 39.0& 39.4\\
\hline  
Re-Sign\cite{koller2017re}& $\surd$& -& 27.1& 26.8& 25.7& 26.6& -& -\\
\hline  
DNF\cite{cui2019deep}& -& $\surd$& 23.8& 24.4& -& -& 32.8& 32.4\\
\hline  
Joint-SLRT\cite{camgoz2020sign}& $\surd$& -& -& -& 24.6& 24.5& 33.1& 33.2\\
\hline  
FCN\cite{cheng2020fully}& $\surd$& -& 23.7& 23.9& 23.3& 25.1& 33.2& 33.5\\
\hline  
VAC\cite{min2021visual}& $\surd$& -& -& 21.2& 22.3 -& -& -& -\\
\hline  
SEN\cite{hu2023self}& $\surd$& -& 19.5& 21.0& 19.3& 20.7& 31.1& 30.7\\
\hline  
STMC\cite{zhou2020spatial}& -& $\surd$& 21.1& 20.7& 19.6& 21.0& -& -\\
\hline  
C2SLR\cite{zuo2022c2slr}& -& $\surd$& 20.5& 20.4& 20.2& 20.4& 31.9& 31.0\\
\hline  
STENet\cite{yin2023spatial}& $\surd$& -& 19.3& 20.3& 19.4& 21.1& 28.9& 28.9\\
\hline  
HST-GNN\cite{kan2022sign}& -& $\surd$& 19.5& 19.8& 20.1& 20.3& -& -\\
\hline  
CorrNet\cite{hu2023continuous}& $\surd$& -& 18.8& 19.4& 18.9& 20.5& 30.6& 30.1\\
\hline  
CorrNet+ACDR\cite{guo2023conditional}& $\surd$& -& 18.6& 19.0& 18.3& 20.0& 29.6& 29.0\\
\hline  
TwoStream-SLR\cite{chen2022two}& -& $\surd$& 18.4& 18.8& 17.7& 19.3& 25.4& 25.3\\
\hline  
MAM-FSD& $\surd$& -& 19.2& 18.8& 18.2& 19.4& 25.8& 24.5\\
\hline  
\end{tabular}
\end{table*}

\begin{figure*}
\begin{minipage}{0.32\textwidth}
\includegraphics[width=2.2in]{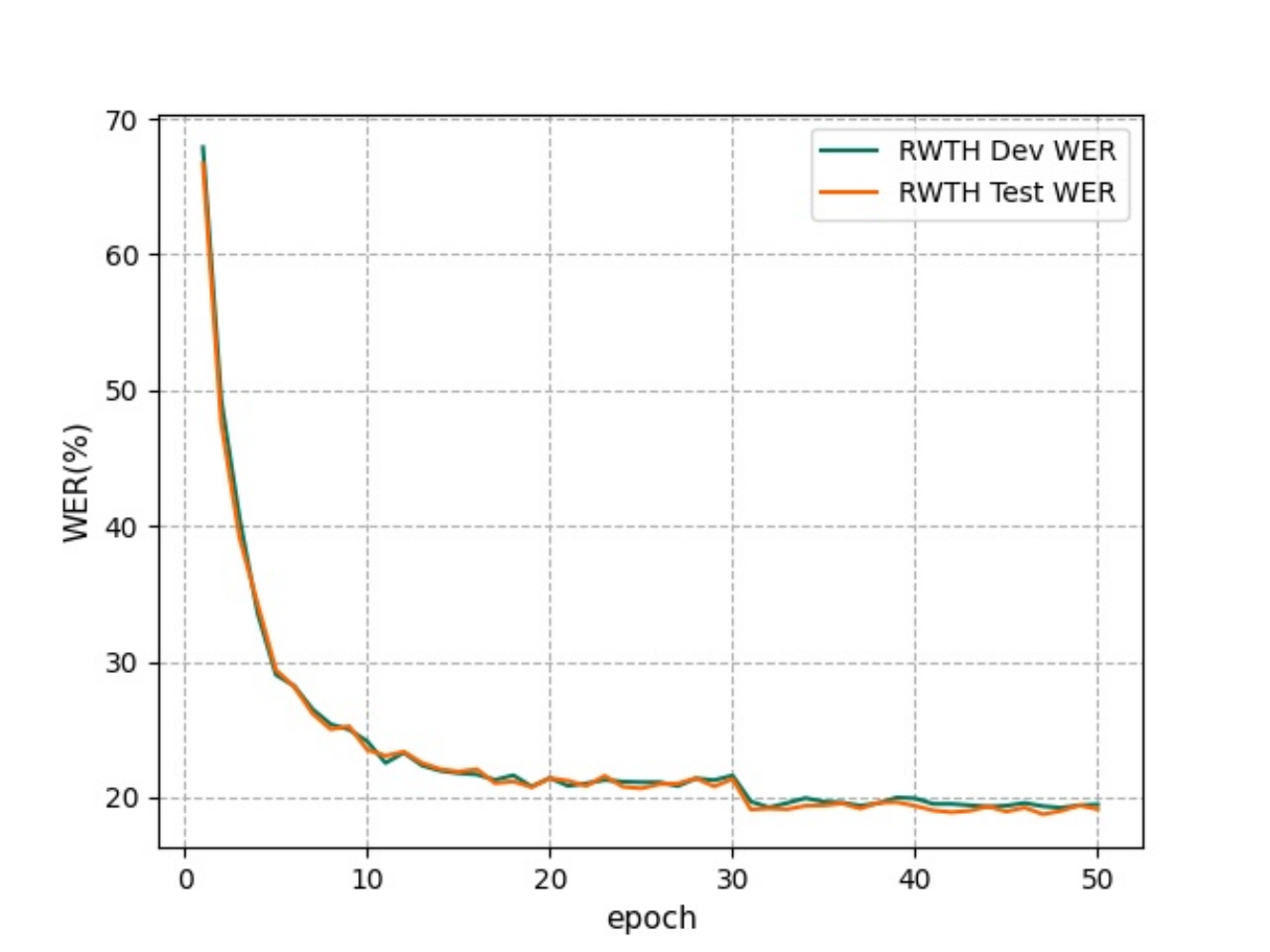}
\caption{WER variation curves for RWTH validation set and test set.}
\label{fig:ljxy4}
\end{minipage}
\hfill
\begin{minipage}{0.32\textwidth}
\includegraphics[width=2.2in]{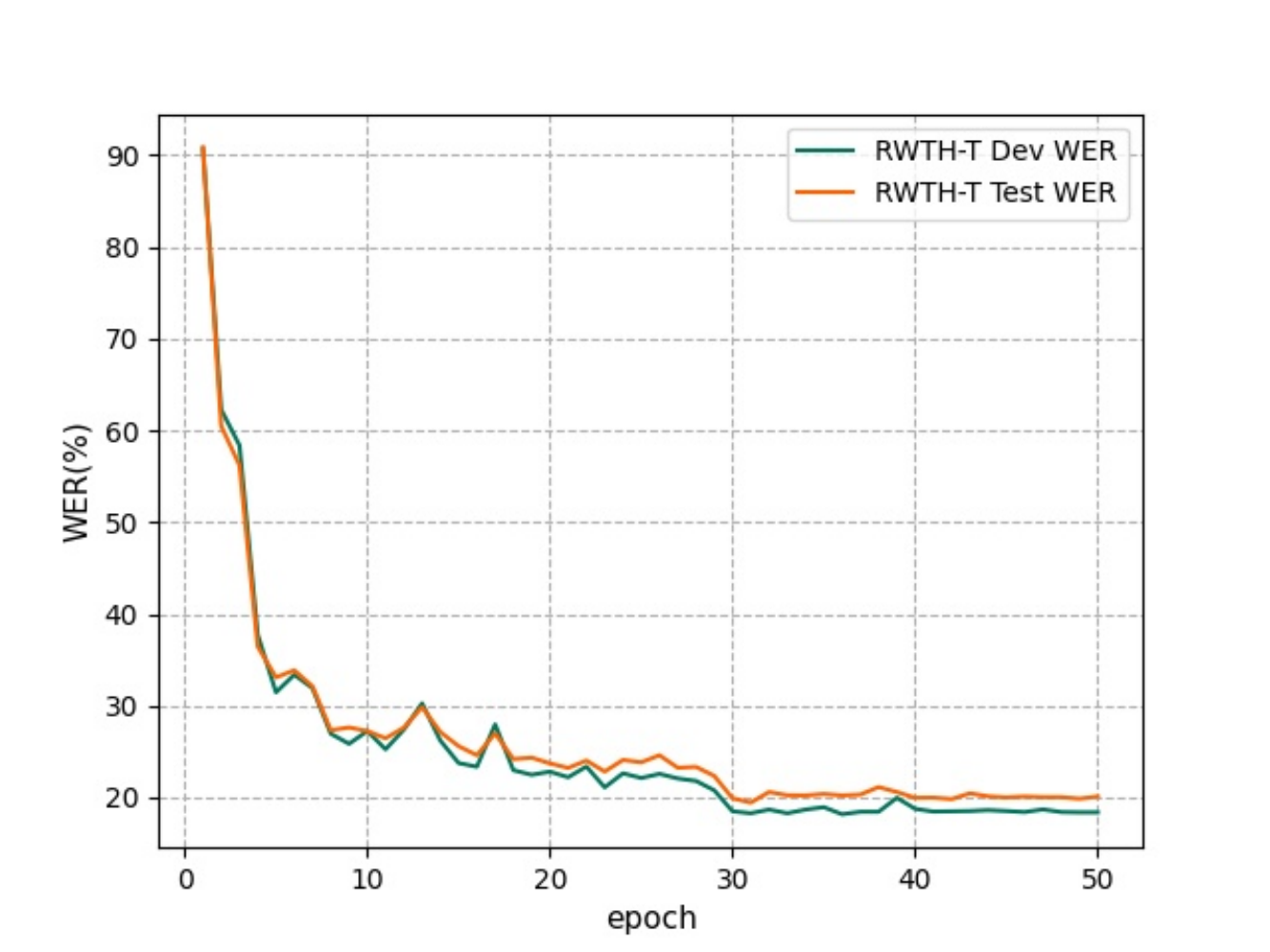}
\caption{WER variation curves for RWTH-T validation set and test set.}
\label{fig:ljxy5}
\end{minipage}
\hfill
\begin{minipage}{0.32\textwidth}
\includegraphics[width=2.2in]{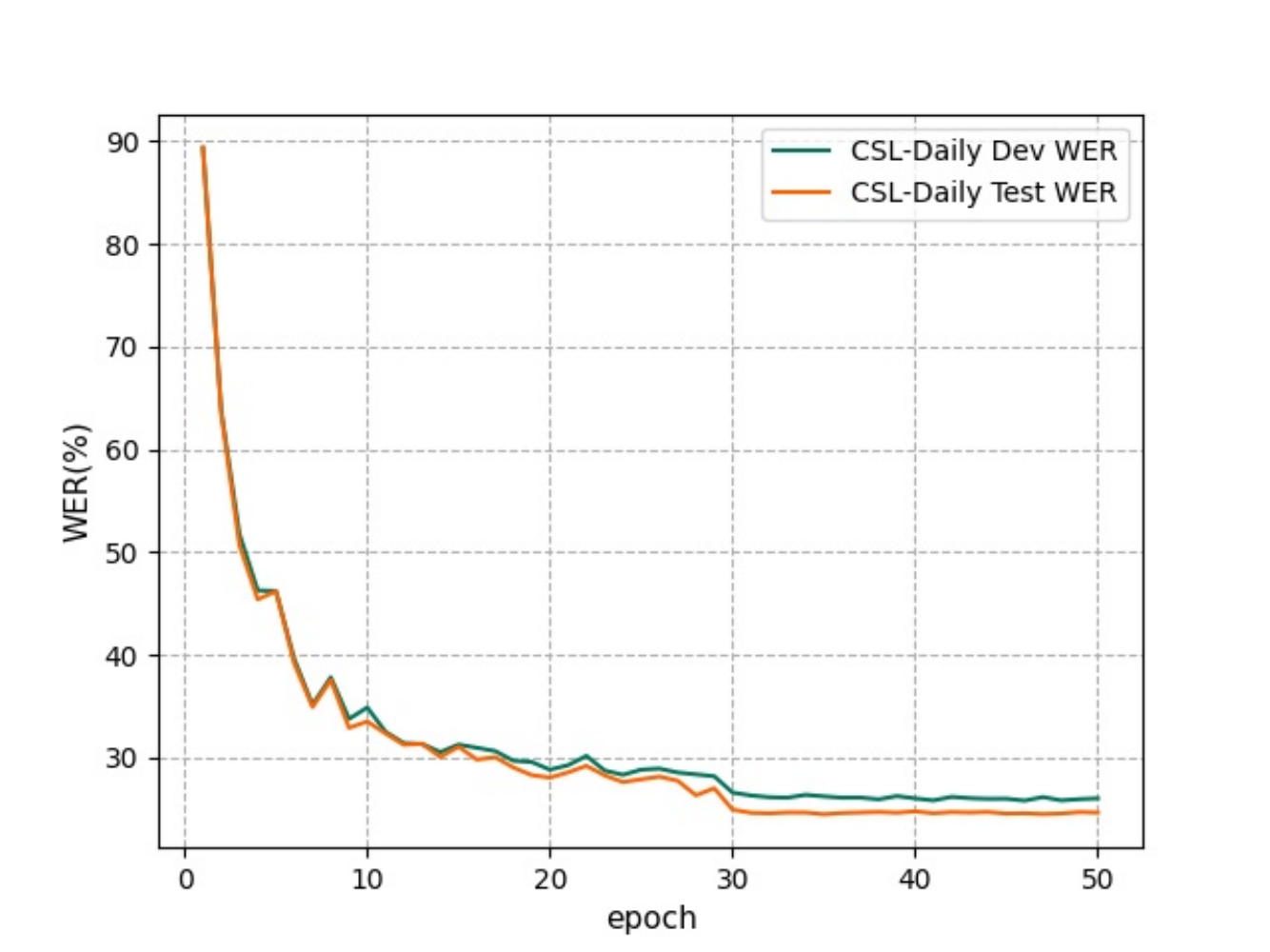}
\caption{WER variation curves for CSL-Daily validation set and test set.}
\label{fig:ljxy6}
\end{minipage}
\end{figure*}

During the training period, for each stage, we use the feature output of that stage as the teacher features, the D-block processed features as the student features, the features after stage 1 are the lowest dimensional features in the frame-level features and are only used as the student features, and the features after stage 4 are the highest dimensional features in the frame-level features and are only used as the teacher features. When performing the self-distillation training, we use the MSE loss function as the self-distillation loss function with the following expression:\par

\begin{equation}
MSE(y,y^{'}) = \frac{\sum_{\substack{i=1}}^{\substack{n}}(y_i-y_i^{'})^{2}}{n} 
\end{equation}

\noindent where $n$ is the number of features, $y_i$ is the i-th teacher feature, and $y_i^{'}$ is the i-th student feature. We performed a total of 3 feature self-distillations and weighted and summed the 3 self-distillation loss functions to get the final self-distillation loss function as follows:\par

\begin{equation}
Loss_{MSE}=\alpha Loss_{MSE_1} + \beta Loss_{MSE_2} + \lambda Loss_{MSE_3}
\end{equation}

\noindent where $Loss_{MSE}$ is the total self-distillation loss, $Loss_{MSE_n}$ is each sub-self-distillation loss, and $n=1,2,3$. $\alpha$, $\beta$, $\lambda$ are the weights, giving each sub-distillation loss a different weight. The sum of this loss function and the final VAE loss constitutes the final training loss function.\par

\section{Experiment}

\subsection {Dataset and Judgment Criteria}

In this paper, we conduct experiments on three large-scale publicly available datasets, RWTH\cite{koller2015continuous}, RWTH-T\cite{camgoz2018neural}, and CSL-Daily\cite{zhou2021improving}. The RWTH dataset is a sign language video recorded by the German weather broadcasting television station, consisting of a total of 6841 different videos. The official training set, validation set, and test set have been divided into 5672, 540, and 629, respectively. The RWTH-T dataset is an extension of the RWTH dataset, which has been expanded in size to include 1085 vocabulary for CSLR tasks. All videos are divided into 7096, 519, and 642 videos for training, validation, and testing. CSL Daily is a newly released large-scale Chinese sign language dataset with an annotated vocabulary of 2000 and a Chinese text vocabulary of 2343. It consists of 18401, 1077, and 1176 samples from the training, validation, and testing sets. We use Word Error Rate(WER)\cite{koller2015continuous} as the evaluation criterion, which is widely used in CSLR. It is the sum of the minimum insertion, replacement, and deletion operations required to convert the recognition sequence into a standard reference sequence, and a lower WER means better recognition performance, as defined below:\par

\begin{equation}
WER=100\%\times \frac{ins+del+sub}{sum}
\end{equation}

Where $ins$ represents the number of words to be inserted, $del$ represents the number of words to be deleted, $sub$ represents the number of words to be replaced, and $sum$  represents the total number of words in the label.\par

\begin{figure*}
  \begin{center}
  \includegraphics[width=6.5in]{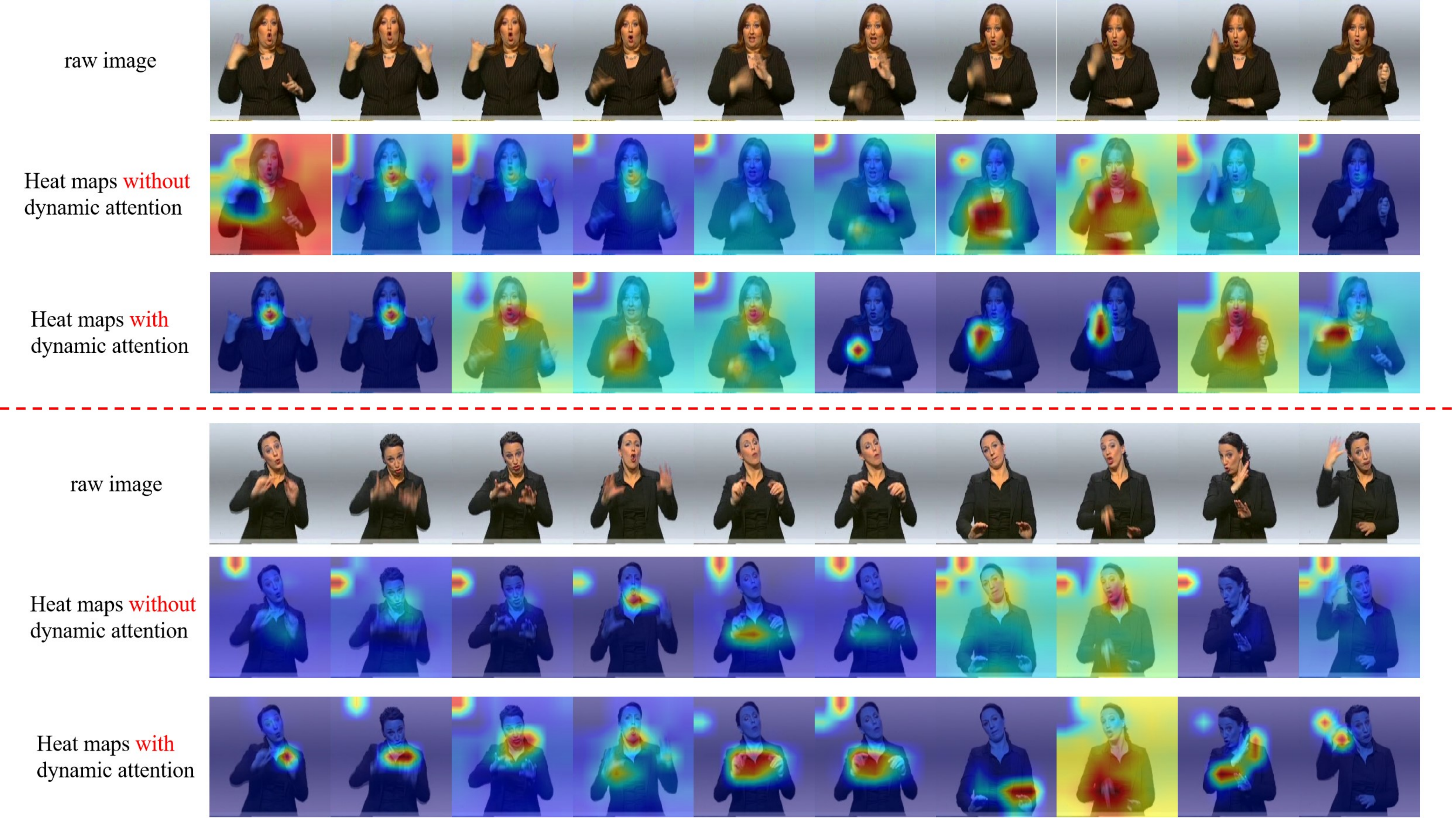}\\
  \caption{Heat map for visualization of the motor attention mechanism module. We visualize the recognition process of two sign language videos from the RWTH dataset by Grad-CAM. Top: original image; middle: heat map without dynamic attention, bottom: heat map with dynamic attention. The heat map shows that our dynamic attention module focuses more on the motion regions in the images, which overlap with the main motion parts of the sign language demonstration actions(regions such as hands and face(red regions)).}\label{fig:ljxy7}
  \end{center}
\end{figure*}

\subsection {Implementation Rules}

For our experimental implementation, the overall model was trained using the Adam optimizer\cite{kingma2014adam}, with the initial learning rate and weight factor set to $10^{-4}$ and a batch size of 2. We added the dynamic attention module and frame-level self-distillation method to the base model VAC, and the frame-level feature extraction backbone was used with ResNet-34. The graphics card used for the experiments was an RTX3090Ti, and the GPU dedicated memory size is 24 G. Data enhancement is performed using random cropping, and random flipping while training on three publicly available datasets, RWTH, RWTH-T, and CSL-Daily. For random cropping, the input data size is $256\times 256$ and the size after random cropping is $224\times 224$. For random flipping, the probability of flipping is set to 0.5. The flipping and cropping processes are performed for the video sequences. In addition, temporal enhancement processing was performed to randomly grow or shorten the length of the video sequences within $\pm 20\%$, and 3 layers of frame-level self-distillation were used for training. A total of 50 epochs were used in the training phase, and the learning rate was reduced by 80\% at the 30th and 40th epochs. The model was tested using only center cropping for data enhancement, and a beam search algorithm was used for decoding in the final CTC decoding stage with a beam width of 10.\par

\begin{table*}[!htbp]
\centering
\caption{Ablation study of the number of dynamic attention modules}
\label{tab:aStrangeTable2}
\begin{tabular}{c|c|c|c|c|c|c}
\hline  
Number of layers in dynamic attention& 0& 1& 2& 3& 4& 5\\
\hline  
Dev(\%)& 19.8& 20.1& 19.4& 19.3& 19.2& 19.0\\
\hline  
Test(\%)& 20.1& 20.1& 19.8& 19.2& 18.8& 19.5\\
\hline  
\end{tabular}
\end{table*}

\begin{table*}[!htbp]
\centering
\caption{Ablation study of the number of convolutional layers in a dynamic attention structure}
\label{tab:aStrangeTable3}
\begin{tabular}{c|c|c|c|c}
\hline  
Number of convolutional layers in dynamic attention structures& 2& 3& 4& 5\\
\hline  
Dev(\%)& 19.4& 18.9& 19.2& 19.2\\
\hline  
Test(\%)& 19.5& 19.2& 18.8& 19.2\\
\hline  
\end{tabular}
\end{table*}

\begin{table*}[!htbp]
\centering
\caption{Ablation study of convolution kernel size in dynamic attention to structural forces}
\label{tab:aStrangeTable4}
\begin{tabular}{c|c|c|c}
\hline  
The size of convolution kernels in dynamic attention structures& 3& 5& 7\\
\hline  
Dev(\%)& 19.2& 18.8& 18.7\\
\hline  
Test(\%)& 18.8& 19.4& 19.2\\
\hline  
\end{tabular}
\end{table*}

\begin{table*}[!htbp]
\centering
\caption{Ablation study of a BN layer in dynamic attention structures}
\label{tab:aStrangeTable5}
\begin{tabular}{c|c|c}
\hline  
BN layer& -& $\surd$\\
\hline  
Dev(\%)& 19.2& 19.3\\
\hline  
Test(\%)& 18.8& 19.3\\
\hline  
\end{tabular}
\end{table*}

\begin{table*}[!htbp]
\centering
\caption{Ablation study with different loss combinations}
\label{tab:aStrangeTable6}
\begin{tabular}{c|c|c|c|c|c|c}
\hline  
Serial number& $loss_{VAE}$& $loss_{MSE_1}$& $loss_{MSE_2}$& $loss_{MSE_3}$& Dev(\%)& Test(\%)\\
\hline  
1& $\surd$& -& -& -& 19.1& 19.5\\
\hline  
2& $\surd$& $\surd$& -& -& 19.9& 19.3\\
\hline  
3& $\surd$& -& $\surd$& -& 19.3& 19.4\\
\hline  
4& $\surd$& -& -& $\surd$& 19.0& 19.2\\
\hline  
5& $\surd$& $\surd$& $\surd$& -& 19.1& 19.2\\
\hline  
6& $\surd$& -& $\surd$& $\surd$& 19.1& 19.4\\
\hline  
7& $\surd$& $\surd$& -& $\surd$& 18.7& 19.4\\
\hline  
8& $\surd$& $\surd$& $\surd$& $\surd$& 19.2& 18.8\\
\hline  
\end{tabular}
\end{table*}

\subsection {Experimental Results}

In this paper, the MAM-FSD is experimented on RWTH, RWTH-T and CSL-Daily respectively, and the model recognition accuracy is shown in Table 1, and the curves generated from the WERs in Table 1 are shown in Figures 4, 5, and 6.\par

From Table 1, it can be seen that MAM-FSD achieves state-of-the-art recognition accuracy compared to other state-of-the-art models for single cue using only RGB images as input. For the experimental results on the RWTH dataset, the WER values reaches 19.2\% and 18.8\% on the validation and test sets respectively; compared to the multi-cue method TwoStream-SLR\cite{chen2022two}, the WER values on the test set are equal; and compared to the single-cue method CorrNet+ACDR\cite{guo2023conditional}, the WER on the test set is reduced by 0.2\%. On the RWTH-T dataset, MAM-FSD achieves WER values of 18.2\% and 19.4\% on the validation and test sets respectively, which is only 0.1\% higher than the WER of the TwoStream-SLR method and 0.6\% lower than that of the CorrNet+ACDR method.The WER values of MAM-FSD on the validation and test sets of CSL-Daily reaches 25.8\% and 24.5\%, which is 0.8\% lower than the WER of the TwoStream-SLR method on the test set. As can be seen in Figures 4, 5, and 6, for the three datasets, the trend of WER values is consistent, all of them are decreasing with epoch increase in WER, and there is a significant decrease in WER at the first lr change, which proves the effectiveness of our model.\par

\subsection {Visualization of Motor attention mechanism}

We conduct visualization experiments on the motor attention mechanism module to more directly demonstrate the value of our proposed module. The visualization experiment is conducted on the RWTH dataset.\par

Figure 7 shows the heat maps for the process of recognizing two sign language videos with and without the addition of the dynamic attention module. We use the Grad-CAM method\cite{selvaraju2017grad} for heat map visualization. In Figure 7, the heat map display is divided into two parts, top and bottom, by red dashed lines, each of which is a separate heat map display. For each video heat map, the top of the image sequence is the original image, the middle is the heat map without dynamic attention, and the bottom is the heat map with dynamic attention. In the heat map, the red area represents the key focus area, and the blue area represents the non key focus area.\par

The heat map comparison shows that with the dynamic attention module than without the module for the image of the hands and face and other regions of greater attention, more focused on the above regions, and the main movement areas in the sign language demonstration process is in the hands and face and other regions. Therefore, the dynamic attention module can make the model focus its attention on the motion region of the image during the recognition process to get the dynamic feature expression that helps inference, which is a very important condition in improving the accuracy of continuous sign language recognition.\par

\subsection {Ablation Experiment}

In this section, we conduct ablation experiments on the RWTH dataset to further validate the effectiveness of each component of the model, in which WER is used as a metric, with smaller WER representing better performance.\par

\textbf{The ablation experiment of the number of motor attention mechanism modules.} We add different numbers of motor attention mechanism modules to the backbone network, and the accuracy comparison results are shown in Table 2. The modules are added sequentially from the input of the network. From Table 2, it can be seen that as the number of motor attention mechanism modules increases, the WER on the test set decreases. The WER reaches a minimum value of 18.8\% when the number of modules is 4. As the number reaches 5, the WER value increases by 0.7\%, indicating a decrease in the accuracy of the model. It can be seen that when the motor attention mechanism module is 4, the model performance is optimal, and the accuracy is improved by 6.5\% on the test set compared to the module without motor attention mechanism, which proves the effectiveness of the motor attention mechanism module.\par

\textbf{The ablation experiment of convolutional layers in motor attention mechanism.} The motor attention mechanism consists of multiple 3D convolutions, which only train adjacent data in the temporal dimension. The number of layers in 3D convolution determines the size of the receptive field of the motor attention mechanism in the temporal dimension, and on the other hand, the increase in the number of convolution layers also increases the nonlinearity of the model. When the number of 3D convolution layers changes, the impact on model accuracy is shown in Table 3. The 3D convolution used in this paper has a fixed kernel size of  . From Table 3, it can be seen that as the number of 3D convolution layers increases, the model accuracy is improving and the highest sign language recognition accuracy is achieved when the number of layers is 4, at which time the WER value on the test set is 18.8\%.\par

\textbf{The ablation experiment of convolutional kernel size in motor attention mechanism.} For multiple 3D convolutions in the motor attention mechanism, the size of the convolution kernel is positively correlated with the size of the receptive field in the motor attention mechanism. The model accuracy obtained by different 3D convolution kernel sizes is shown in Table 4. From Table 4, it can be seen that as the size of the convolution kernel increases, the model accuracy decreases. When the size of the convolution kernel is 3, the model accuracy is highest. After that, when the size of the convolution kernel increases to 5 and 7, the model accuracy on the test set decreases by 3.2\% and 2.1\%, respectively, compared to the convolution kernel size of 3.\par

\textbf{The ablation experiment of BN layer in dynamic attention structure.} Generally speaking, adding a BN layer after a convolutional layer is helpful for model performance. This is because the BN layer can change the data distribution, effectively control the gradient explosion to prevent the gradient from disappearing, and at the same time can accelerate the convergence of the model to prevent the appearance of overfitting. However, the motor attention mechanism focuses on the motion changes within adjacent time domains, and adding a BN layer may not necessarily have a positive effect. We add a BN layer to the convolutional layer in the motor attention mechanism, and a comparison of the experimental results with and without the BN layer is shown in Table 5. From Table 5, it can be seen that without adding a BN layer after each convolutional layer, directly connecting with the activation function has the best effect, and its WER value reaches the minimum value on both the validation and test sets.\par

\textbf{The ablation experiment of different loss combinations.} There are several different losses in the MAM-FSD model, which are mainly classified into two categories based on their functions, one is the frame-level self-distillation loss function, which aims to increase the feature expression capability without increasing the computational resources, and the second is the VAE loss function, which aims to establish visual alignment constraints to enhance the alignment supervision of the feature extractor. We combine the different losses and obtain the model accuracy as shown in Table 6. As can be seen from the data in rows 1-4 in Table 6, the combination with any of the frame-level self-distorting losses($loss_{MSE_n}$,$n=1,2,3$) reduces the WER values on the test set, which is a good illustration of the effectiveness of $loss_{MSE}$. The combination of losses from rows 1, 2, 5, and 8 shows that the WER on the test set is decreasing when the number of $loss_{MSE}$ is increased on $loss_{VAE}$. The best performance is achieved when the model is trained using all the losses and the WER reaches a minimum value of 18.8\%, which is a 3.6\% improvement in accuracy over the model obtained using only $loss_{VAE}$.\par

\section{Conclusion}

In this paper, we propose a motor attention mechanism module to obtain a dynamic feature representation of an image by studying frame-to-frame variations and capturing the distortion changes in localized motion regions when generating a sign language expression, allowing the model to focus more on the dynamic semantic knowledge that contributes to inference. Visualization shows that our motor attention mechanism module focuses more on motion regions in continuous sign language videos, which overlap with regions such as hand and face regions that are mainly associated with sign language actions. We also apply self-distillation method for the first time in frame-level feature extraction for continuous sign language, which in combination with the motor attention mechanism module constitutes the novel CSLR model proposed, MAM-FSD, that expresses superiority over other state-of-the-art models on three publicly available datasets.\par 

\section*{Acknowledgment}

This work was supported in part by the Development Project of Ship Situational Intelligent Awareness System, China under Grant MC-201920-X01, in part by the National Natural Science Foundation of China under Grant 61673129. \par

~\\\par
\textbf{Data availability} The datasets used in the paper are cited properly.\par

\section*{Declarations}

\textbf{Conflict of interest} The authors declare that they have no known competing financial interests or personal relationships that could have appeared to influence the work reported in this paper.\par


%





\ifCLASSOPTIONcaptionsoff
  \newpage
\fi





\bibliographystyle{IEEEtran}
\bibliography{IEEEabrv,Bibliography}

\vfill


\end{document}